\documentclass[runningheads]{llncs}
\usepackage{graphicx}
\usepackage{amsmath}
\usepackage{amsfonts} 

\usepackage{booktabs}
\usepackage{geometry}
\usepackage{subcaption}
\usepackage{hyperref}

\begin{document}
\title{ Recurrence With Correlation Network for Medical Image Registration}
%
%
\author{Vignesh Sivan\inst{1} \and
Teodora Vujovic\inst{2} \and
Raj Ranabhat\inst{2} 
\and Alexander Wong \inst {1} 
\and Stewart Mclachlin \inst{1} \and
Michael Hardisty \inst{2} }
\authorrunning{V. Sivan et al.}
%
\institute{University of Waterloo, 200 University Ave W, Waterloo, ON, Canada\and
Sunnybrook Research Institute, 2075 Bayview Ave., Toronto, ON, Canada}
%
\maketitle              
\begin{abstract}
We present \textit{Recurrence with Correlation Network} (RWCNet), a medical image registration network with multi-scale features and a cost volume layer. We demonstrate that these architectural features improve medical image registration accuracy in two image registration datasets prepared for the MICCAI 2022 Learn2Reg Workshop Challenge. On the large-displacement National Lung Screening Test (NLST) dataset, RWCNet is able to achieve a total registration error (TRE) of 2.11mm between corresponding keypoints without finetuning. On the OASIS brain MRI dataset, RWCNet is able to achieve an average dice overlap of 81.7\% for 35 different anatomical labels. It outperforms another multi-scale network, the Laplacian Image Registration Network (LapIRN), on both datasets. Ablation experiments are performed to highlight the contribution of the various architectural features. While multi-scale features  improved validation accuracy for both datasets, the cost volume layer and number of recurrent steps only improved performance on the NLST dataset. This result suggests that cost volume layer and iterative refinement using RNN provide good support for optimization and generalization in large-displacement medical image registration. The code for RWCNet is available at \href{https://github.com/vigsivan/optimization-based-registration}{https://github.com/vigsivan/optimization-based-registration}.

\end{abstract}

\section{Introduction}
\label{sec:introduction}

Medical image registration describes the task of spatially aligning one medical image to another. It has a myriad of uses, including image-guided pre-operative planning \cite{risholm_multi-modal_2011} and atlas creation for population-level studies \cite{toga_role_2001}.  Traditionally, image registration has been solved using optimization methods. There is growing interest in the application of deep learning to the task of medical image registration. An advantage of using deep learning over conventional optimization approaches is that it can often arrive at optimal solutions orders of magnitude faster. Based on deep learning's success in the field of computer vision, it may also yield more accurate and more robust spatial transforms than conventional optimization \cite{balakrishnan_voxelmorph_2019}. Indeed, recent work on deep learning image registration (DLIR) has shown its potential for yielding faster and more accurate transforms than conventional methods for some datasets \cite{balakrishnan_voxelmorph_2019,heinrich_voxelmorph_2022,mok_large_2020}.

In terms of inputs and outputs, medical image registration is closely related to optical flow, where the objective of the latter is to compute a flow field describing the motion of objects between two images of, typically, the same scene. Optical flow is an important and well-studied computer vision problem and it has a range of applications including action recognition and pose estimation. Common architectural features of optical flow networks include multi-scale features \cite{sun_pwc-net_2018} and cost volume layers \cite{ilg_flownet_2016,sun_pwc-net_2018,teed_raft_2020}. 

The computation of multi-scale features and cost volumes for 3D medical image registration have been separately explored in prior work \cite{mok_large_2020,heinrich_closing_2019}, which have demonstrated improvements in registration performance on standard datasets. This paper introduces a novel, optical flow-inspired network architecture for DLIR, \textit{recurrence with correlation network} (RWCNet). RWCNet combines multi-scale iterative features and a cost-volume layer for medical image registration. To the authors' knowledge, no other work combines these architectural features for DLIR. The contributions of this paper are listed as follows:



\begin{itemize}
    \item We present a novel network architecture for DLIR, \textit{recurrence with correlation network} (RWCNet), for DLIR that outperforms other continuous-domain and multiscale networks in some standard registration datasets prepared for the MICCAI 2022 Learn2Reg Workshop.
    \item Ablations of architectural components are performed to demonstrate the performance of the various architectural features of the registration network. The results indicate that the contributions of various architectural features can vary by dataset.
    
\end{itemize}

\section{Related Work}

\subsubsection{Voxelmorph}

Voxelmorph is one of the earliest and best known methods for DLIR \cite{balakrishnan_voxelmorph_2019}. It trains a 3D UNet \cite{cicek_3d_2016} to learn a sub-voxel displacement field, $D$, that jointly optimizes image fidelity of the transformed image, denoted $\mathcal{S}$, and a regularization loss promoting smooth spatial transformation, denoted $\mathcal{R}$. The loss function, $L$, describing the goodness of fit for registering a moving image $m$ to a fixed image $f$ is expressed as:

\begin{equation}
    \label{eq:vxm}
    L = \mathcal{S}(m \circ (Id + D), f) + \mathcal{R}(D)
\end{equation}

where $Id$ is the identity transform and $m \circ (Id + D)$ resamples the moving image onto a new grid parametrized by the displacement field, $D$. This resampling operation is carried out differentiably using the spatial transformer network (STN) \cite{jaderberg_spatial_2016}. The loss function described by \eqref{eq:vxm} can be augmented with the correspondence of auxiliary data such as segmentations and keypoints. Voxelmorph was found to be competitive with optimization-based registration methods such as Demons \cite{thirion_non-rigid_1996} and Symmetric Normalization \cite{avants_symmetric_2008}.

\subsubsection{Laplacian Image Registration Network}

Laplacian Pyramid Image Registration Network (LapIRN) \cite{mok_large_2020} learns flow fields at $N=3$ resolutions. At each resolution a CNN with residual skip connections learns a displacement field by jointly optimizing an image fidelity term and smoothness term described by \eqref{eq:vxm}. LapIRN is trained in a coarse to fine manner; thus the CNN's at low resolutions are trained before training the networks responsible for learning at higher resolutions. The flow fields learned at low resolutions are upsampled and used to warp the moving input image at higher resolutions. 

Multi-scale refinement operates on the principle that an optimal displacement field at low resolution is also a good displacement field at high resolution \cite{mok_large_2020}. Moreover, at coarse resolution the optimization problem is simpler, owing to the fact that the required displacements are smaller in magnitude and there are fewer high-level features to match; refining the flow field from a coarse-to-fine resolution can thus simplify the optimization problem. LAPIRN showed improvement in the large-displacement setting, when compared to Voxelmorph and conventional approaches. LapIRN uses convolutional neural network (CNN) architecture with residual connections to learn the flow fields at each resolution.

The method presented in this paper also learns flow fields from coarse to fine resolutions, like LapIRN. RWCNet, the architecture presented in this paper, differs from the one used by LapIRN; a recurrent CNN architecture with features encoders and a cost volume layer is used.

\subsubsection{RAFT}

The architecture of RWCNet is directly inspired by RAFT, which achieves impressive performance in optical flow \cite{teed_raft_2020}. The RAFT network architecture has three key characteristics:  1) CNN feature encoders, 2) cost volume computation at multiple scales between all pairs and 3) recurrent CNN architecture for computing and refining the flow field and performing a `lookup' that subsamples the global cost volume. RWCNet differs from RAFT in several ways. First, computing the cost between all pairs of voxels becomes prohibitively expensive for 3D volumes. Furthermore, computing the cost volume at multiple resolutions simultaneously is also computationally expensive. As such, RWCNet consists of three sub-networks that learn displacement fields at three different resolutions, similar to LapIRN for DLIR or PWCNet\cite{sun_pwc-net_2018} for optical flow.




\section{Methods}

\subsubsection{Sub-network Architecture.} The architecture for the recurrent CNN sub-network is shown in Figure 1. Given a fixed and moving image pair ($f$ and $m$, respectively), the network first learns fixed and moving features ($F_f$ and $F_{m,0}$) by feeding both images through a feature extractor network. A voxel-wise correlation between the fixed and moving features is computed, $C_0$. Due to the large number of dimensions, the correlation is restricted so that only voxels within a certain range, $r$ of the moving voxel are considered. Additionally, the moving image is fed through a context network that extracts contextual information for the hidden network. The output of the contextual network is used as the initial hidden state of the RNN, $h_0$. Finally, a displacement field with zero displacement, $D_0$ is initialized.

The hidden state, flow, cost volume and moving image are fed into an update block that is a modified gated recurrent unit (GRU) \cite{cho_properties_2014}. The GRU is almost identical in implementation to the one used by RAFT \cite{teed_raft_2020}, and it outputs a new hidden state and a new displacement field, $h_1$ and $\Delta D$. The new displacement field is used to update the aggregate displacement, i.e, $D_1 = \Delta D + D_0$. This new displacement field is used to warp the moving features (generating $F_{m,1}$), which can be used to generate a new cost volume, $C_1$, for the next RNN time step. This process of updating displacement field with GRU cell and generating cost volume is repeated for $N$ RNN time steps.

\begin{figure}
\includegraphics[width=\textwidth]{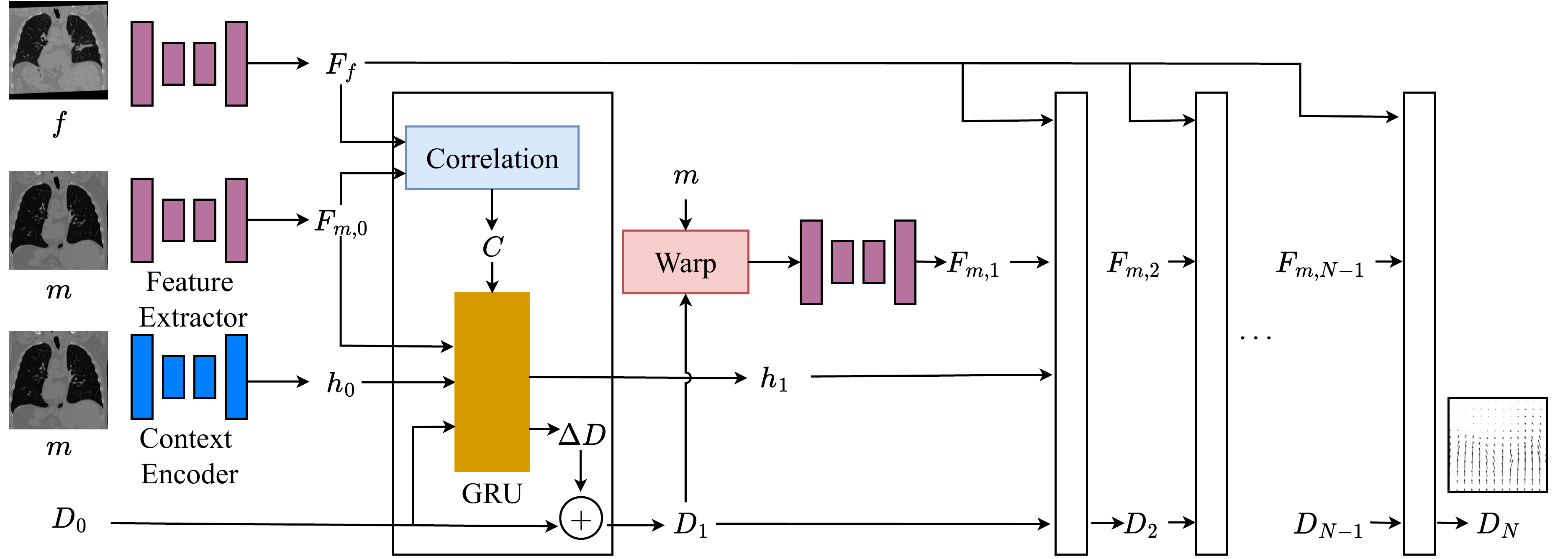}
\caption[RWCNet sub-network architecture]{RNN Sub-network architecture. It has 3 main components: 1) feature extractor and context encoder networks for extracting fixed and moving features as well as a context 2) computation of a cost volume through correlation of the input features 3) update aggregate displacement using a GRU-based update block. } \label{fig1}
\end{figure}

\subsubsection{Coarse to Fine Registration.} We adopt a course to fine approach to image registration. For each resolution, $s$, a new RNN is trained using inputs from the previous resolution. The weights from previous resolutions are frozen at finer resolutions.  At fine resolutions computing the cost volume for the whole volumes becomes prohibitively expensive; as such, our approach divides the input images into uniform, non-overlapping windows or patches. The size of the patches at each resolution is parameterized by the `patch factor', $p^s \in [0,1]$. The size of the patches at resolution $s$ is computed as $p^s \times S$ where $S$ is the size of the full image at $1\times$ resolution. At higher resolutions, the flow field is used to warp the initial moving image (or patch) using flow fields computed at lower resolutions. Furthermore, the final hidden state is cached at lower resolutions and added to the initial hidden state at higher resolutions, increasing the non-linearity of the network and providing additional context to the network.

\begin{figure*}
\includegraphics[width=\textwidth]{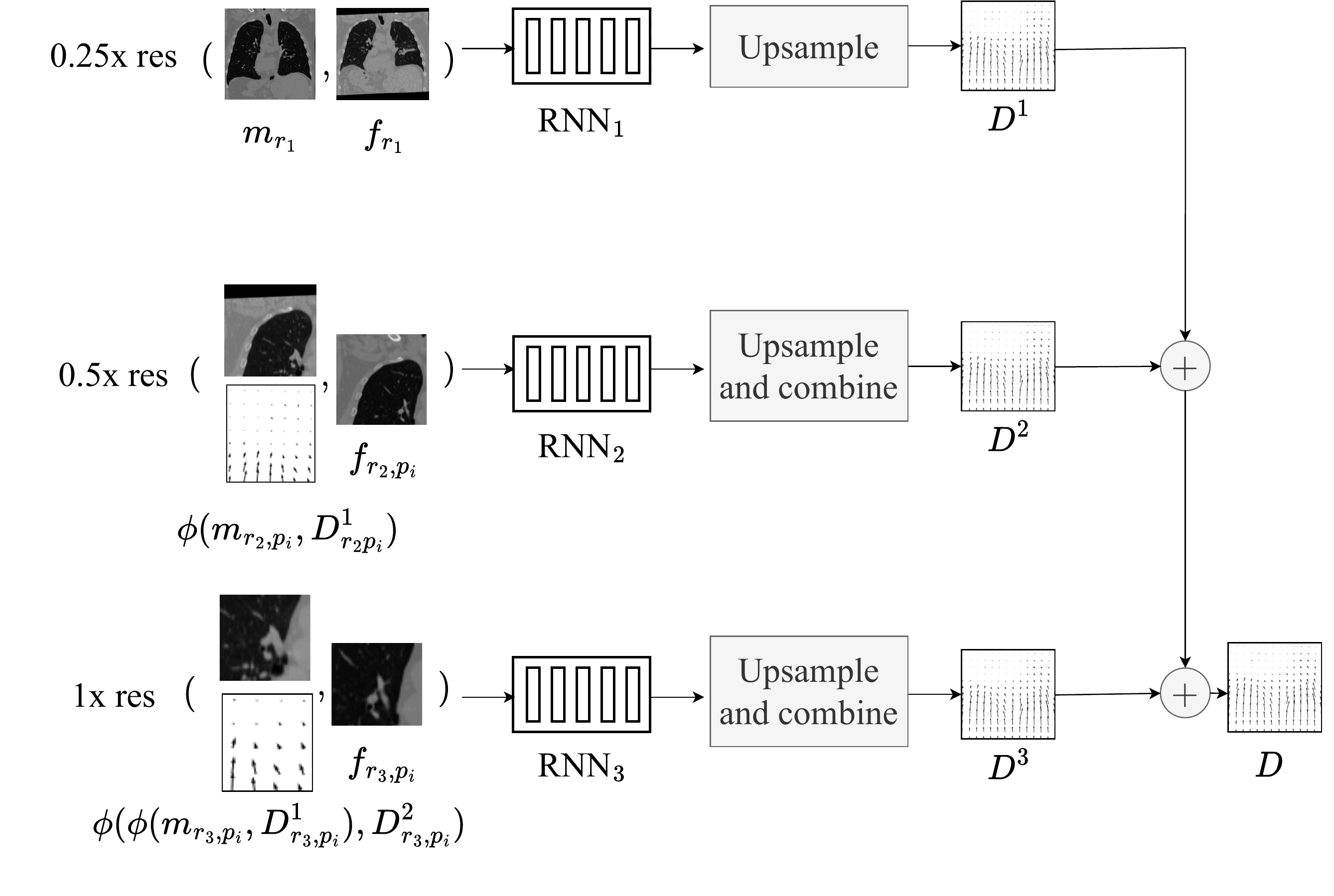}
\caption[RWC Multi-scale refinement with finetuning at inference time]{ RWCNet: Multi-scale network, showing refinement of flow field using separate RNN sub networks(\ref{fig:fig1}) for each resolution. Note, at higher resolutions, patches of the input image are fed into the network, rather than the entire image. Not shown is the fact the final hidden layer is upsampled and concatenated with the first hidden state of the next resolution.} \label{fig:global}
\end{figure*}

\subsubsection{Ablation Experiments.} Ablation experiments are performed to assess the contribution of the individual architectural features. To test the impact of multi-scale refinement, the network is trained to learn registration fields at a single resolution. To study the impact of correlation, the cost volume computation is removed; instead the input to the GRU is simply a concatenation of the input features, which is consistent with Voxelmorph and LapIRN. 




\section{Experiments}

\subsection{Datasets}

Experiments are performed with the OASIS \cite{marcus_open_2007} and NLST \cite{national_lung_screening_trial_research_team_national_2011} datasets prepared for the MICCAI 2022 Learn2Reg workshop challenge \cite{noauthor_learn2reg_nodate}. The OASIS dataset consists of 414 T1-weighted MRI scans of individuals from ages 18-96 with mild to severe Alzheimer's. The scans are skull-stripped and resampled onto an isotropic grid and cropped to a uniform size. 35 segmentation labels are provided for important brain regions. The dataset is split into 395 images for training and 19 for validation. Intersubject registration in this context could be used for constructing a sub-population brain atlas or for analysing intensity changes in consistent brain regions that are linked to disease progression. 

NLST is a lung-CT dataset with pairs of inhale/exhale scans; keypoints and masks are provided by the Learn2Reg challenge for semi-supervised training. We use a subset of the image pairs (100 out of 150) of the NLST dataset released by the Learn2Reg challenge for training and validation, with a 90:10 training/validation. Since respiration is accompanied by a large change in lung volume, the displacement field required to register NLST is large, relative to OASIS.

\subsection{Training Parameters}

The subnetworks at each scale were trained separately from coarse to fine. At higher resolutions, corresponding patches of the fixed and moving images are passed into the network to decrease GPU memory requirements. Table 1 shows the number of steps and the sizes of the inputs at each resolution. To address overfitting, dropout with probability 0.5 is used for the feature networks. The network takes about 30 hours to train both  the NLST and OASIS datasets on an NVIDIA A100 with 32GB of RAM.

For OASIS, the similarity component of the loss function was a summation of the mean squared error (MSE) between the warped moving image and the fixed image intensities as well as the Dice loss between the warped segmentation and the fixed segmentation. The regularization loss was the average gradient of the flow field, as used in Voxelmorph. For NLST, the data was range normalized to between 0 and 1, with -4000 serving as the minimum value and 16000 serving as the maximum value. The loss function was a weighted summation of the MSE and the total registration error (TRE). The TRE measures the discrepancy in mm between corresponding keypoints in the fixed and moving images. The mean gradient of the flow field was used as the regularization loss on the flow field. For both datasets, an Adam optimizer with learning rate of $3\times10^{-4}$ was used.

LAPIRN, another multi-resolution model, was trained on OASIS using training parameters from \cite{mok_large_2020}. We use the non-diffeomorphic variant, which does not ensure topological consistency of the spatial transformation, but achieves greater quantitative accuracy. For NLST, we augment training with a supervised discrepancy loss between the fixed and moving keypoints once the displacement is applied to the moving keypoints.  We use a MSE loss instead of the normalized cross correlation (NCC) loss prescribed by the original paper to be consistent with the the loss used for training RWCNet. 

\begin{table}[]
\centering
\caption{Resolution-specific Parameters.}
\label{tab1}
\begin{tabular}{@{}ccccc@{}}
\toprule
Resolution & RNN Steps & Patches Per Image & Patch Factor & Training Steps          \\ \midrule
 0.25       & 12        & 1       & $0.25$ & 30000        \\
 0.5        & 12        & 8       & $0.25$ & 45000        \\
 1          & 4         & 8       & $0.5$ & 60000       \\ \bottomrule

\end{tabular}
\end{table}

\section{Results} \label{sec:3res}

\begin{figure}
  \centering
  \begin{tabular}{@{}c@{}}
    \includegraphics[]{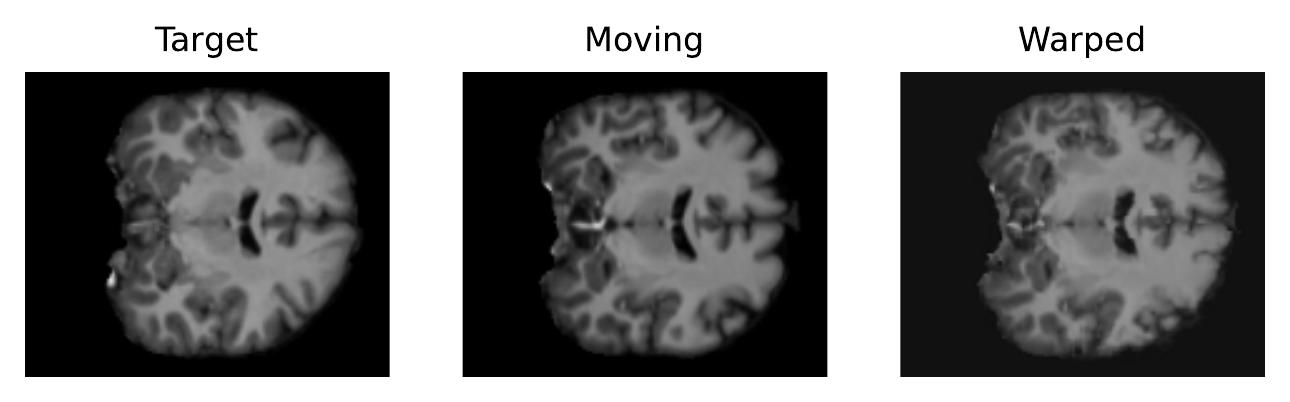} \\[\abovecaptionskip]
    \small (a) OASIS sample results
  \end{tabular}

  \vspace{\floatsep}

  \begin{tabular}{@{}c@{}}
    \includegraphics[]{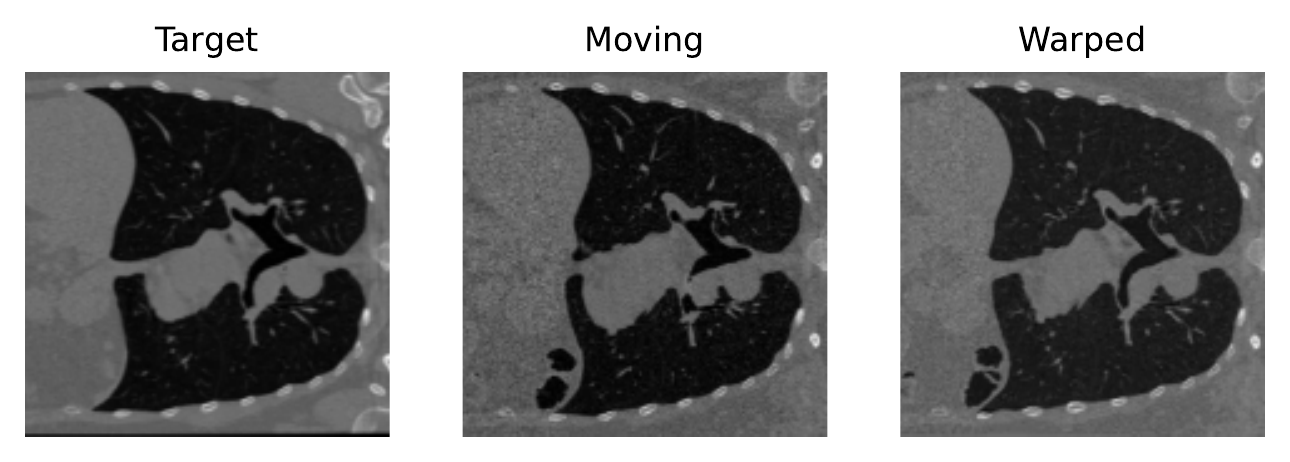} \\[\abovecaptionskip]
    \small (b) NLST sample results
  \end{tabular}

  \caption[Flow fields for OASIS and NLST Datasets]{Qualitative results, showing the target, moving and warped images for OASIS (a) and NLST (b).}\label{fig:disps}
\end{figure}

Figure \ref{fig:disps} shows qualitative results generated for the NLST and OASIS datasets. Table \ref{tab:results} summarizes the results when comparing RWCNet with LAPIRN on the NLST and OASIS datasets. RWCNet outperforms LAPIRN on both datasets. However, the difference in Dice is only 0.7\%, which might be entirely explainable by random weight initialization in network training. For the NLST dataset, the difference in performance is much more pronounced, with the difference in average TRE being $>$3mm. These results suggest that the architectural features of RWCNet, significantly aid generalization performance in the more challenging large-displacement setting.

\begin{table}[]
\caption{Experiment Results on NLST and OASIS Validation}
\label{tab:results}
\begin{center}
\begin{tabular}{@{}ccc@{}}
\toprule

Experiment & NLST TRE (mm) $\downarrow$ & 
OASIS Dice (\%) $\uparrow$\\ \midrule

Zero Displacement & 9.73 & 52.4 \\
LAPIRN \cite{mok_large_2020}    & 5.51 & 80.0 \\ 
\textbf{RWCNet}      & \textbf{2.11} & \textbf{80.7} \\

 \bottomrule
\end{tabular}
\end{center}
\end{table}

Table \ref{tab:ablation} shows results from the architectural ablation tests. These ablation results \ref{tab:ablation} provide interesting insights into the role that architecture plays in registration accuracy in different datasets. Unsurprisingly, multi-resolution registration plays a crucial role in the accuracy of RWCNet; registering at only $4\times$ downsampling on the NLST dataset yields a keypoint discrepancy of 5.52 mm, whereas registering at multiple resolutions yields a discrepancy of 2.11mm. OASIS Dice, likewise, drops from 80.7\% to 74.0\%.

\begin{table}[]
\caption{Ablation Experiment Results on NLST and OASIS Datasets}
\label{tab:ablation}
\begin{center}
\begin{tabular}{@{}ccc@{}}
\toprule

Experiment & NLST TRE (mm) $\downarrow$ & 
OASIS Dice (\%) $\uparrow$\\ \midrule

RWCNet     & 2.11 & 80.7 \\
RWCNet with single resolution (4x) & 5.52 & 74 \\
RWCNet without correlation      & 4.10 & 80.0 \\
RWCNet with 2-timestep GRU     & 5.17 & 80.1 \\

 \bottomrule
\end{tabular}
\end{center}
\end{table}

The impact of correlation and number of RNN time steps is markedly different for both datasets. In the OASIS datasets, replacing correlation with stacking of the input feature tensors does not drastically impact the registration performance. The Dice score drops by 0.7. Likewise, when only 2 RNN time steps are used in RWCNet, the drop in accuracy is even lower in the OASIS dataset; the Dice score only drops by 0.6\%.  This is in contrast to NLST, where decreasing the number of RNN time steps and removing correlation drastically decrease performance. The keypoint accuracy decreases to 4.10mm when correlation is not computed. Likewise, when only 2 time steps are used, the accuracy decreases to 5.17mm. 

The varying impact of ablation between the 2 datasets may be explained by the nature of the datasets.  OASIS has relatively small displacements compared to NLST and as such may not benefit as much from cost volumes and the RNN structure. This observation could be helpful for designing generalizable methods to large-displacement datasets.




\section{Conclusion}


The investigation developed RWCNet for medical image registration, showing good performance 2 open datasets.  RWCNet includes architectural features common in optical flow, a multi-scale approach, explicit cost volume computation, and iterative refinement with a RNN.  The optical flow features were found to be most useful in the NLST dataset, with large displacements and had little impact on the OASIS dataset results.   Future work should investigate how the performance of these architectural features changes for other datasets to confirm the importance of these features for large deformation problems. This can inform future work towards developing dataset-dependent self-configuring registration methods.

\bibliographystyle{ieeetr}
\bibliography{references}





\end{document}